\documentclass[twocolumn,final,times]{elsarticle}

\usepackage{lineno,hyperref}
\modulolinenumbers[5]
\usepackage{framed,multirow}
\usepackage{amsmath}
\usepackage{mathabx}
\usepackage{pgfplots}
\usepackage{booktabs}
\usepackage{array}
\usepackage{xcolor}
\usepackage{graphicx}
\usepackage{subcaption}
\usepackage{color}
\usepackage{bm}
\usepackage{rotating}

\newcolumntype{C}[1]{>{\centering\let\newline\\\arraybackslash\hspace{0pt}}m{#1}}

\usepackage{lineno,hyperref}

\journal{Pattern Recognition Letters}

\biboptions{authoryear}

\begin{document}

\begin{frontmatter}

\title{Dual Rectified Linear Units (DReLUs): A Replacement for \emph{Tanh} Activation Functions in Quasi-Recurrent Neural Networks}

\author[ugent]{Fr\'{e}deric Godin}
\author[ugent]{Jonas Degrave}
\author[ugent]{Joni Dambre}
\author[ugent,korea]{Wesley De Neve}

\address[ugent]{IDLab, ELIS, Ghent University -- imec, Ghent, Belgium}
\address[korea]{Center for Biotech Data Science, Ghent University Global Campus, Incheon, Korea}

\begin{abstract}

In this paper, we introduce a novel type of Rectified Linear Unit (ReLU), called a Dual Rectified Linear Unit (DReLU). A DReLU, which comes with an unbounded positive and negative image, can be used as a drop-in replacement for a \emph{tanh} activation function in the recurrent step of Quasi-Recurrent Neural Networks (QRNNs) \citep{DBLP:journals/corr/BradburyMXS16}. Similar to ReLUs, DReLUs are less prone to the vanishing gradient problem, they are noise robust, and they induce sparse activations.

We independently reproduce the QRNN experiments of \cite{DBLP:journals/corr/BradburyMXS16} and compare our DReLU-based QRNNs with the original \emph{tanh}-based QRNNs and Long Short-Term Memory networks (LSTMs) on sentiment classification and word-level language modeling.
Additionally, we evaluate on character-level language modeling, showing that we are able to stack up to eight QRNN layers with DReLUs, thus making it possible to improve the current state-of-the-art in character-level language modeling over shallow architectures based on LSTMs.

\end{abstract}


\end{frontmatter}


\section{Introduction}
Rectified activation functions are widely used in modern neural networks. They are more commonly known as Rectified Linear Units (ReLUs), which are essentially neurons with a rectified activation function \citep{icml2010_NairH10}. ReLUs are part of many successful feed-forward neural network architectures such as ResNets \citep{he2016deep} and DenseNets \citep{dense}, reaching state-of-the-art results in visual recognition tasks. Compared to traditional activation functions such as \emph{sigmoid} and \emph{tanh}, ReLUs offer a number of advantages: (1) they are simple and fast to execute, (2) they mitigate the vanishing gradient problem, and (3) they induce sparseness. As a result, ReLUs permit training neural networks with up to 1000 layers \citep{he2016deep}.

However, due to their unboundedness, rectified activation functions have not experienced the same success in Recurrent Neural Networks (RNNs), compared to the \emph{sigmoid} and \emph{tanh} activation functions. 
The recently introduced Quasi-Recurrent Neural Network (QRNN) \citep{DBLP:journals/corr/BradburyMXS16}, a hybrid recurrent/convolutional neural network, can partly solve this issue by avoiding hidden-to-hidden matrix multiplications. However, using a rectified activation function instead of a \emph{tanh} activation function to calculate a new hidden state is still cumbersome. Indeed, given that ReLUs only have a positive image, strictly negative values cannot be added to a hidden state, thus making the subtraction of values impossible.

In this paper, we present a new type of neural network component, called a Dual Rectified Linear Unit (DReLU). Rather than having a unit with a single activation function, DReLUs subtract the output of two regular ReLUs, thus coming with both a positive and negative image. Consequently, similar to ReLUs, DReLUs make it possible to avoid vanishing gradients and to enforce sparseness. Moreover, similar to \emph{tanh} units, DReLUs have a negative image. However, different from \emph{tanh} units, DReLUs can be \emph{exactly} zero. Indeed, \emph{tanh} units have difficulties being exactly zero, often leading to the introduction of noise. As a result, we can state that DReLUs exhibit properties of both ReLUs and \emph{tanh} units, thus making it possible for DReLUs to replace the functionality of neurons that make use of a \emph{tanh} activation function, while at the same time preventing the occurrence of vanishing gradients when stacking multiple layers. Finally, we also introduce a variant based on the recently introduced Exponential Linear Units (ELUs) \citep{clevert2015fast}, called Dual Exponential Linear Units (DELUs).

We extensively evaluate DELUs and DReLUs on three different natural language processing tasks, comparing with the orginal QRNN architecture \citep{DBLP:journals/corr/BradburyMXS16}: (1) sentiment classification, (2) word-level language modeling, and (3) character-level language modeling.


Our main contributions can be summarized as follows:
\begin{itemize}
\item We introduce a new neural network component, called Dual Rectified Linear Unit (DReLU). This newly proposed unit is able to successfully replace units with a tanh activation function in Quasi-Recurrent Neural Networks (QRNNs). 
\item We independently reproduce the QRNNs experiments of \cite{DBLP:journals/corr/BradburyMXS16} on sentiment classification and word-level language modeling. Additionally, we introduce a third evaluation task, namely character-level language modeling. To sustain reproducibility and transparency, the full implementation of all three models is available online \citep{godinimpl}.

\item We demonstrate that we can easily stack up to eight DReLU-based QRNN layers without the need of skip connections, obtaining state-of-the-art results in character-level language modeling.

\end{itemize}

The remainder of the paper is organized as follows. In Section~\ref{sec:relatedwork}, we give an overview of related work. This overview is followed by a more formal problem definition and solution in Section~\ref{sec:methodology}. In Section~\ref{sec:experiments}, we evaluate the proposed solution by means of four different benchmarks. Finally, we end with conclusions in Section~\ref{sec:conclusions}.



\section{Related Work}\label{sec:relatedwork}
The term Rectified Linear Unit (ReLU) was coined by \cite{icml2010_NairH10}. A ReLU is a neuron or unit with a rectified linear activation function. A rectified linear activation function is defined as follows:
\begin{equation}
\label{eq:relu}
f_{ReL}=max(0,x).
\end{equation}
One particular issue with ReLUs is that the image of the negative domain of the ReLU is always zero. Therefore, the gradient will not flow through that neuron during backpropagation. This led to the introduction of Parametric ReLUs \citep{DBLP:conf/iccv/HeZRS15}, which multiply input values in the negative domain with a small value, rather than setting it to zero. However, the non-saturating behavior in the negative domain can introduce noise, while the image of standard ReLUs is always exactly zero. More recently, Exponential Linear Units (ELUs) \citep{clevert2015fast} showed significant improvements over ReLUs. While ELUs sacrifice the simplicity of ReLUs, they have a natural way of combating the bias shift and at the same time being noise robust. An exponential linear activation function is defined as follows:
\begin{equation}
\label{eq:elu}
f_{EL}(x) = 
\begin{cases}
x & \quad if \quad x > 0 \\
\alpha\ (exp(x)-1) & \quad if \quad x  \leqslant 0 \\
\end{cases}
\end{equation}
with  $\alpha > 0$. The parameter $\alpha$ controls the saturation value of the image for the negative domain. 

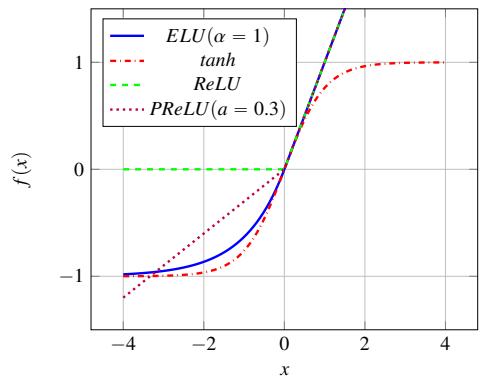
\begin{figure}[t]
\centering
\begin{tikzpicture}[scale=0.75]
\begin{axis}[domain=-4:4,ymin=-1.5,ymax=1.5, samples=100,grid=major,xlabel=$x$,ylabel=$f(x)$, legend pos=north west]
\addplot [blue, solid, very thick]   {(x<0) * (exp(x)-1)   +
     (x>=0) * (x)};
\addplot [ red, dashdotted, very thick]    {tanh(x)};
\addplot [green,dashed, very thick]    {max(0,x)};
\addplot [purple, dotted, very thick] {max(0.3*x,x)};

\legend{$ELU (\alpha=1)$ ,$tanh$, $ReLU$, $PReLU (a=0.3)$ }
\end{axis}
\end{tikzpicture}
\caption{Visualization of four important activation functions.}
\label{fig:relu_overview}
\end{figure}



Despite the success of ReLUs in Convolutional Neural Networks (CNNs), ReLUs have not experienced the same success in Recurrent Neural Networks (RNNs). Due to the application of a recurrent step on a possibly infinite sequence, bounded activation functions such as \emph{sigmoid} or \emph{tanh} are needed. \cite{DBLP:journals/corr/LeJH15} managed to train a simple RNN with a ReLU activation function by carefully initializing the hidden state weight matrix with the identity matrix. Though good results are obtained, they are still outperformed by Long Short-Term Memory networks (LSTMs) \citep{Hochreiter:1997:LSM:1246443.1246450}. In general, ReLUs are used in feed-forward parts of neural networks, rather than in recurrent parts \citep{DBLP:journals/corr/CollinsSS16}. 

Due to the variable length of text, RNNs are still the preferred type of neural network for modeling sequences of words, and LSTMs in particular \citep{Schmidhuber201585}. 
CNNs are less commonly used because they need large receptive fields of fixed-input size to capture long-range dependencies. Only very recently, a number of hybrid approaches have emerged, also leveraging CNNs. \cite{DBLP:journals/corr/KalchbrennerESO16} used a fixed-size CNN to perform character-level neural machine translation, obtaining state-of-the-art results. \cite{DBLP:journals/corr/BradburyMXS16} introduced the concept of Quasi-Recurrent Neural Networks (QRNNs). The two main components of QRNNs are CNNs and an LSTM-based simplified recurrent step. The simplified recurrent step allows for modeling sequences of variable length without using a fixed-size window, while the combination with CNNs effectuates significant speed-ups. 

All current methods for modeling variable-length sequences still rely on the \emph{tanh} activation function. This activation function, however, causes gradients to vanish during backpropagation. ReLUs are less prone to the vanishing gradient problem but are unbounded and only have a positive image. Hence, they are less suited for RNNs. The proposed concept of a DReLU leverages the strengths of both types of units in order to train models on variable-length sequences.

\section{Methodology}\label{sec:methodology}
In this section, we first argue why ReLUs cause training issues in Recurrent Neural Networks (RNNs) such as simple RNNs \citep{elman} and Long Short-Term Memory networks (LSTMs) \citep{Hochreiter:1997:LSM:1246443.1246450}. Next, we explain Quasi-Recurrent Neural Networks (QRNNs) \citep{DBLP:journals/corr/BradburyMXS16}, which have proven to be a viable alternative to LSTMs.
Finally, we introduce the concept of DReLUs, as well as the exponential extension, termed DELUs.

\subsection{Motivation}
RNNs are neural networks that contain cyclic connections. These cyclic connections allow preserving history and are well suited for modeling variable-length sequences. There exist many types of RNNs, most notably simple RNNs and LSTMs. 

In a simple RNN \citep{elman}, the hidden state at time step $t$ is defined as follows:
\begin{equation}
\label{eq:rnnhiddenstate}
\bm{h_t} = g(W\cdot\bm{h}_{t-1}+U\cdot\bm{x}_t+\bm{b}),
\end{equation}
in which $W$ and $U$ are weight matrices and $\bm{b}$ is a bias vector. The function $g$ is the activation function, which is in practice a \emph{sigmoid} or \emph{tanh} function. The functions \emph{sigmoid} and \emph{tanh} are bounded functions. This implies that they are functions for which there exists a real number $M$ such that $|g(x)| \leq M$ for all $x$ in domain $X$. Both the activation functions \emph{sigmoid} and \emph{tanh} are bounded by $1$. Consequently, all hidden states are also bounded and their absolute value is never larger than $1$.

If we simplify the recurrent step defined in Equation~\ref{eq:rnnhiddenstate} to 
\begin{equation}
\bm{h}_t = g(W\cdot\bm{h}_{t-1}),
\end{equation}
then
\begin{equation}
\bm{h}_t = g(W\cdot g(W\cdot ... g(W\cdot\bm{h}_0))).
\end{equation}
If $g$ is an unbounded activation function such as a ReLU, it becomes clear that the hidden state can grow exponentially large. Indeed, if the largest eigenvalue of the weight matrix is larger than one, then the norm of the vector $\bm{h}_t$ will continuously grow and eventually explode \citep{Goodfellow-et-al-2016}.


The same holds true for LSTMs for which both the current cell state and hidden state are dependent on the previous hidden state via matrix multiplications. Consequently, replacing the \emph{tanh} activation function with an unbounded activation function would still be problematic, leading to a similar unbounded multiplication as with simple RNNs. 

\subsection{Quasi-recurrent neural networks}
QRNNs are hybrid neural networks inspired by LSTMs, bringing together the advantages of CNNs and LSTMs \citep{DBLP:journals/corr/BradburyMXS16}. CNNs are fast and highly parallelizable while LSTMs are able to naturally model long-range dependencies. However, both simple RNNs and LSTMs contain a slow recurrent step (cf. Eq.~\ref{eq:rnnhiddenstate}), which can only be executed in sequence. Indeed, $\bm{h}_t$ depends on $\bm{h}_{t-1}$, which depends in its turn on $\bm{h}_{t-2}$, and so on, thus creating an execution bottleneck.



To reduce the computational effort needed for the recurrent step, the hidden-to-hidden matrix multiplications of an LSTM were removed and a convolution of size $n$ over the input $\bm{x}$ was introduced. Additionally, the input and forget gates were connected, yielding only two gates, namely a forget gate $\bm{f}_t$ and an output gate $\bm{o}_t$. More formally, a QRNN with fo-pooling is defined as follows:
\begin{gather}
\bm{f}_t, \bm{o}_t  = \sigma ( U_{f,o} \cdot \bm{x}_{t-n+1:t}),\\
\label{eq:candidate}
\bm{\tilde{c}_t} = tanh(U_{c} \cdot \bm{x}_{t-n+1:t}).\\
\label{eq:qrnn_ct_fo}
\bm{c}_t = \bm{c}_{t-1} \odot \bm{f}_t+\tilde{\bm{c}_t} \odot (\bm{1}-\bm{f}_t).\\
\bm{h}_t = \bm{c}_{t} \odot \bm{o}_t.
\end{gather}

Consequently, the recurrent step is reduced to a simple gated/weighted sum, thus effectuating faster execution.
This opens up possibilities for unbounded activation functions such as rectified linear activation functions, given that the hidden state cannot exponentially grow anymore. 

\subsection{Dual rectified linear units}
As introduced in the previous section, QRNNs do not contain hidden-to-hidden matrix multiplications anymore, thus avoiding hidden state explosions when an unbounded activation function is used. However, replacing the \emph{tanh} function of the candidate cell state $\tilde{\bm{c}_t}$ with a ReLU (cf. Eq.~\ref{eq:relu}) would limit the expressiveness of a cell state update (cf. Eq.~\ref{eq:qrnn_ct_fo}). Indeed, the values of the gates are bounded between zero and one. If the activation function of the candidate cell state $\tilde{\bm{c}_t}$ is a ReLU, then the cell state can only be updated with positive values. Consequently, this means that only values can be \emph{added} to the previous cell state $\bm{c}_{t-1}$, but not \emph{subtracted}. As a result, the only way to lower the values of the hidden state is to set the gates to zero. The \emph{tanh} function does not have this issue, given that its image is bounded between minus one and one. Therefore, the cell state can also be updated with negative values.

To be able to update the cell state with negative values and, at the same time, use the unbounded rectified linear activation function, we introduce the novel concept of a Dual Rectified Linear Unit (DReLU): rather than using a single ReLU, we propose to subtract two ReLUs.


The dual rectified linear activation function is a two-dimensional activation function. In analogy with the rectified linear activation function, the proposed function can be formally defined as follows: 
\begin{equation}
f_{DReL}(a,b) = max(0,a)-max(0,b)
\end{equation}
or
\begin{equation}
f_{DReL}(a,b) = 
\begin{cases}
0 & \quad if \quad a \leq 0 \quad and \quad b \leq 0 \\
a & \quad if \quad a > 0 \quad and \quad b \leq 0 \\
-b & \quad if \quad a \leq 0 \quad and \quad b > 0 \\
a - b & \quad if \quad a > 0 \quad and \quad b > 0 \\
\end{cases}
\end{equation}
in which $a$ and $b$ are scalar values. A DReLU can be seen as being a unit with three separate states. It can be exactly zero, negative or positive.

An important benefit of DReLUs over \emph{tanh} activation functions is the ability to be exactly zero. Indeed, sparseness allows efficient training of larger stacks of neural networks \citep{pmlr-v15-glorot11a} and makes (D)ReLUs more noise-robust compared to other related activation functions such as Leaky ReLUs.

The partial derivatives of the dual rectified linear activation function with respect to $a$ and $b$ are similar to the standard rectified linear activation function:
\begin{equation}
\frac{\partial f_{DReL}}{\partial a} = 
\begin{cases}
0 & \quad if \quad a \leq 0  \\
1 & \quad if \quad a > 0  \\
\end{cases}
\end{equation}
and
\begin{equation}
\frac{\partial f_{DReL}}{\partial b} = 
\begin{cases}
0 & \quad if \quad b \leq 0  \\
-1 & \quad if \quad b > 0 \\
\end{cases}
\end{equation}
Different from the \emph{sigmoid} and \emph{tanh} activation functions, DReLUs are less prone to vanishing gradients, a property also shared with ReLUs. When a ReLU or DReLU is active, the magnitude of the gradient through the activation function is neither amplified nor diminished. Additionally, rectified linear activation functions are fast to execute. 

Finally, we can replace the calculation of the candidate cell state $\tilde{\bm{c}}_t$ of a QRNN, which uses a \emph{tanh} activation function (cf. Eq.~\ref{eq:candidate}), with a DReLU:
\begin{equation}
\tilde{\bm{c}}_t = max(0,U_{c,1} \cdot \bm{x}_{t-n+1:t})-max(0,U_{c,2} \cdot \bm{x}_{t-n+1:t}).
\end{equation}


\subsection{Dual exponential linear units}
ELUs have shown to perform better than ReLUs for image classification \citep{clevert2015fast}. Indeed, the non-zero image of the negative domain combats bias shift and speeds up learning. However, ELUs contain more complex calculations and cannot be exactly zero. 
In analogy with DReLUs, we can define DELUs. A dual exponential linear activation function can be expressed as follows:
\begin{equation}
f_{DEL}(a,b) = f_{EL}(a)-f_{EL}(b)
\end{equation}
in which $f_{EL}$ is defined as in Equation~\ref{eq:elu}. $f_{DEL}$ can be perceived as a smoother version of $f_{DReL}$. Note that although $f_{EL}$ is only zero when the input is zero, $f_{DEL}$ saturates to zero when the inputs $a$ and $b$ go to minus infinity.

\section{Experiments}\label{sec:experiments}
We evaluate the proposed DReLUs and DELUs as part of a Quasi-Recurrent Neural Network on three different sequencing tasks. First, we reproduce the results of \citep{DBLP:journals/corr/BradburyMXS16} for sentiment classification and word-level language modeling, comparing our DReLUs with \emph{tanh}-based QRNNs and LSTMs. Second, we evaluate DReLUs and DELUs on a third task that is more challenging in nature, namely character-level language modeling. The latter consists of two benchmarks.

To support reproducibility, the exact implementation of all experiments can be found online \citep{godinimpl}, including the experiments of the original QRNN paper \citep{DBLP:journals/corr/BradburyMXS16} for which no full implementation is available.

\subsection{Sentiment Classification}
The first task we consider is sentiment classification of IMDb movie reviews. This is a binary classification task of documents with up to 2818 tokens. The goal is to compare our DReLUs and DELUs as part of a QRNN with the original \emph{tanh}-based QRNN model and LSTMs. We also investigate the impact of densely connecting all layers.

\begin{table*}[t]
\centering
\caption{Evaluation of QRNNs with DReLU or DELU activation functions, compared to LSTMs and a QRNNs with \emph{tanh} activation functions for the task of sentiment classification. The (mean) accuracy and standard deviation is reported. DC means Densely Connected.}
\begin{tabular}{ cC{1.3cm}C{1.3cm}cC{1.5cm} }
\toprule
\textbf{Name} & \textbf{Activation function} & \textbf{Hidden state size} & \textbf{\#Params} & \textbf{Accuracy} \\ 
\toprule
 \multicolumn{5}{c}{\textbf{\cite{DBLP:journals/corr/BradburyMXS16}}} \\
\bottomrule
DC-LSTM  & tanh  & 256 & 3.86M & 90.9\\
DC-QRNN  & tanh   & 256 & 4.21M & 91.4\\
\toprule
\multicolumn{5}{c}{\textbf{Our implementation (Stacked)}} \\
\bottomrule
LSTM  & tanh  & 256  & 2.15M & 90.9 $\pm$ 0.2\\
QRNN & tanh  & 300   & 2.17M & 90.5 $\pm$ 0.2\\
\midrule
\multirow{2}{*}{QRNN}  & DReLU  & 256 & 2.19M & \textbf{91.0} $\pm$ 0.2\\ 
 & DELU  & 256 & 2.19M & \textbf{91.0} $\pm$ 0.1\\ 
\toprule
\multicolumn{5}{c}{\textbf{Our implementation (Densely Connected)}} \\
\bottomrule
DC-LSTM  & tanh  & 256  & 3.86M & \textbf{91.2} $\pm$ 0.2 \\
DC-QRNN & tanh & 242 & 3.85M & 91.1 $\pm$ 0.1  \\
\midrule
\multirow{2}{*}{DC-QRNN}   & DReLU & 200 & 3.84M & \textbf{91.2} $\pm$ 0.2 \\ 
& DELU & 200 & 3.84M  & 91.1 $\pm$ 0.2\\ 
\bottomrule
\end{tabular}
\label{tab:tab_sentiment}
\end{table*}

\subsubsection{Experimental setup}
We follow the exact same setup as \cite{DBLP:journals/corr/BradburyMXS16}. All our models are initialized with 300-dimensional GloVe word embeddings \citep{pennington2014glove} and four stacked recurrent layers. To regularize, dropout ($p=0.3$) is applied between every layer and L2 regularization of $4\times10^{-6}$ is used. As no initialization strategy was specified, we empirically found that using Glorot normal initialization yielded similar results as reported by \cite{DBLP:journals/corr/BradburyMXS16}.

The IMDb movie review dataset \citep{maas-EtAl:2011:ACL-HLT2011} contains 25k positive and 25k negative reviews equally divided in a set for training and a set for testing. 
Because no exact validation set is specified, we split the training set in five distinct validation sets and run every experiment five times with a different validation set. While \cite{DBLP:journals/corr/BradburyMXS16} only report experiments with Densely Connected (DC) layers \citep{dense}, we also report results of experiments without DC layers. In this case, densely connecting all layers means adding concatenative skip connections between every layer, except for the final classification layer.

\subsubsection{Discussion}
The mean and standard deviation over five runs for our different experiments is reported in Table~\ref{tab:tab_sentiment}. In general, we can observe only small accuracy differences (0.1\%) between Densely Connected (DC) LSTMs and QRNNs, independent of the activation function used. However, when no skip connections are used, \emph{tanh}-based QRNNs obtain worse accuracy results (90.5) compared to LSTMs and DReLU-based QRNNs (90.9 and 91.0, respectively). Moreover, the absolute difference in accuracy between DReLU-based QRNNs and DC-QRNNs is only 0.2\% while for \emph{tanh}-based QRNNs and DC-QRNNs, this is 0.6\%. Indeed, skip connections are typically used to avoid vanishing gradients. ReLU-based activation functions are less prone to vanishing gradients than \emph{tanh} activation functions, favoring DReLUs over \emph{tanh} activation functions in QRNNs. Similar conclusions apply for DELUs. 

Our single best model was a DReLU-based QRNN with an accuracy of 91.6\%. However, we believe that a direct comparison with the experimental results of \cite{DBLP:journals/corr/BradburyMXS16} is not meaningful because the validation set is unknown.

Additionally, we compared our QRNN with an LSTM and observed a speed-up of 2,5$\times$ and 2,1$\times$, for a DReLU- and \emph{tanh}-based QRNN over an LSTM, respectively. Consequently, DReLU-based QRNNs obtain similar results as LSTMs at more than double the speed, without deploying skip connections as needed by \emph{tanh}-based QRNNs.

\begin{table*}[!t]
\centering
\caption{Evaluation of different activation functions on the task of word-level language modeling of the PTB corpus.}
\begin{tabular}{ ccccc }
\toprule
\textbf{Name} &  \textbf{Activation function} & \textbf{\#Parameters} & \textbf{Valid perplexity} & \textbf{Test perplexity} \\ 
\midrule
LSTM - medium \citep{Zaremba} & tanh & 20M & 86.2 & 82.7\\
QRNN \citep{DBLP:journals/corr/BradburyMXS16} & tanh & 18M & 82.9 & 79.9\\
\midrule
LSTM - medium (our implementation) & tanh & 20M & 85.8 & 82.6 \\
 QRNN (our implementation) & tanh & 18M & 84.9 & 80.0 \\
 \midrule
 \multirow{3}{*}{QRNN (our implementation)}  & ReLU & 18M & 89.5 & 85.3  \\
  & DReLU & 20M & 82.6 & \textbf{78.4} \\ 
  & DELU & 20M & 83.1 & 78.5 \\ 
\bottomrule
\end{tabular}
\label{tab:tab_word}
\end{table*}
\begin{table}[!t]
\centering
\caption{Activation statistics of the cell states $c_t$ in a QRNN for three different activation functions applied to the candidate cell states $\tilde{c}_t$. To account for near-zero results, we use an interval of $]-0.1;0.1[$. The statistics are calculated on the full test set. }
\begin{tabular}{ cccc }
\toprule
&\textbf{Tanh} & \textbf{ReLU} & \textbf{DReLU}  \\ 
\midrule
Nearly zero activations & 10.02\% &	80.22\%	& 53.90\% \\
Negative activations & 45.28\% &	0.00\% & 23.77\% \\
Positive activations &44.70\% &	19.78\% &	22.33\% \\
\bottomrule
\end{tabular}
\label{tab:activations}
\end{table}

\subsection{Word-level language modeling}
The goal of word-level language modeling is to estimate the conditional probabilities $Pr(w_t|w_1,...,w_{t-1})$ by predicting the next word $w_t$ given all previous words $[w_1,...,w_{t-1}]$.

Rather than focusing on obtaining state-of-the-art perplexity results, in this section, we put the emphasis on investigating the relative performance of ReLUs, DReLUs, and DELUs, compared to stacked LSTMs and stacked QRNNs with \emph{tanh} activation functions in similar conditions.

\subsubsection{Experimental setup}
To properly evaluate our proposed units, we compare with a reimplementation of (1) \cite{Zaremba}, which uses LSTMs, and (2) \cite{DBLP:journals/corr/BradburyMXS16}, which uses QRNNs. Both papers evaluate their proposed architecture on the Penn Treebank dataset \citep{ptb} and use the same setup.


We follow the exact setup of \cite{DBLP:journals/corr/BradburyMXS16} for training QRNNs with DReLUs and DELUs. That is, a two-layer stacked QRNN with 640 hidden units equally sized embeddings and a convolution of size two. For the QRNN with a \emph{tanh} activation function, we initialized the weights uniformly in the interval $[-0.05;0.05]$. In QRNNs with rectified activation functions, we initialized the weights randomly, following a normal distribution $N(0;0.1)$. For DELUs, we set $\alpha$ to $0.1$. For training the neural network, we use the same training procedure as described in \cite{DBLP:journals/corr/BradburyMXS16}. 





    
\begin{table*}[t]
\centering
\caption{Comparison of DReLUs and DELUs with other neural network architectures on the Penn Treebank test set.}
\begin{tabular}{ cC{2cm}cC{2cm}cc }
\toprule
\textbf{Model name} & \textbf{Activation function} & \textbf{\# Layers} & \textbf{Hidden state size} & \textbf{Test BPC} & \textbf{\# Params} \\ 
\midrule
HF-MRNN \citep{mikolov2012subword} & & & & 1.41 \\
BatchNorm LSTM \citep{DBLP:journals/corr/CooijmansBLC16} & & & & 1.32 \\
LayerNorm HM-LSTM \citep{chung+al-2017-multiscale-iclr} & & & & 1.24 \\
LSTM \citep{hypernetworks} & tanh & 2 & 1000 & 1.28 & 12.26M\\ 
Layer Norm HyperLSTM \citep{hypernetworks} & tanh & 2 & 1000 & 1.22 & 14.41M\\  
\bottomrule
\multirow{9}{*}{QRNN (our implementation)} 
 & tanh & 4 & 581 & 1.26 & 6.66M\\
  & tanh & 8 & 579 & 1.22 & 14.69M \\
\cline{2-6}
 & DELU & 4 & 500 & 1.25 & 6.66M\\
 & DELU & 8 & 500 & \textbf{1.21} & 14.69M \\
\cline{2-6}
 & DReLU & 2 & 250 & 1.38 & 0.82M\\
 & DReLU & 4 & 250 & 1.30 & 1.83M\\
 & DReLU & 4 & 500 & 1.25 & 6.66M\\
 & DReLU & 8 & 250 & 1.25 & 3.85M\\
 & DReLU & 8 & 500 & \textbf{1.21} & 14.69M \\

\bottomrule
\end{tabular}
\label{tab:tab_char1}
\end{table*}

\begin{table*}[t]
\centering
\caption{Comparison of DReLUs with other neural network architectures on the Hutter Prize test set.}
\begin{tabular}{ cc }
\toprule
\textbf{Model name} &  \textbf{Test BPC} \\ 
\midrule
Layer Norm LSTM \citep{hypernetworks} & 1.40 \\
Layer Norm HyperLSTM \citep{hypernetworks} & 1.34 \\
Layer Norm HM-LSTM \citep{chung+al-2017-multiscale-iclr} & 1.32 \\
ByteNet \citep{DBLP:journals/corr/KalchbrennerESO16} & 1.31 \\
Recurrent Highway Networks \citep{DBLP:journals/corr/ZillySKS16} & 1.27 \\
\midrule
QRNN (DReLU - 4 layers - 1000 units) & 1.32 \\
QRNN (DReLU - 8 layers - 1000 units) & \textbf{1.25} \\
\bottomrule
\end{tabular}
\label{tab:tab_char2}
\end{table*}
\subsubsection{Discussion}

The results of our experiments are shown in Table~\ref{tab:tab_word}. In  general, we can observe that QRNNs with DReLUs or DELUs outperform a QRNN with a \emph{tanh} activation function applied in the candidate cell state $\tilde{\bm{c}_t}$. More interestingly, a QRNN using a single ReLU activation function performs much worse. The perplexity gap between a QRNN with a single and a dual ReLU activation function is 7 perplexity points. Consequently, ReLUs are inferior replacements for \emph{tanh}-based units but DReLUs can successfully replace them and even outperform them.

When comparing with a LSTM, following the same setup as \citep{Zaremba}, we can observe that we outperform a medium-sized LSTM with 650 hidden units.


To analyze the activation pattern the different activation functions induce, we have also calculated the activation statistics of the cell states $\bm{c}_t$ of a QRNN with a \emph{tanh}, ReLU and DReLU activation function. The results are depicted in Table~\ref{tab:activations}. While only 10.02\% of the cell states $\bm{c}_t$ is nearly zero when using a \emph{tanh} activation function, 80.22\% is nearly zero when using a ReLU and 53.9\% when using a DReLU. The non-zero cell states $\bm{c}_t$ for both DReLU and \emph{tanh} activation functions are equally divided between positive and negative values. Moreover, both QRNNs using ReLUs or DReLUs are equally active in the positive domain. Consequently, DReLU activation functions can show similar behavior as \emph{tanh} activation functions but with the additional benefit of being able to be exactly zero and induce sparse activations. Indeed, sparse layer outputs allow for information disentangling and variable-size intermediate representations \citep{pmlr-v15-glorot11a}, and eventually for training large neural networks, in terms of both width and depth.

In this section, we showed that DReLUs and DELUs have similar properties as \emph{tanh} activation functions and that both are suitable replacements. A single ReLU, however, is an inferior replacement for a \emph{tanh} activation function. Additionally, when using DReLUs, the cell state $\bm{c}_t$ becomes sparse, which is an important benefit when training large neural networks.  


\subsection{Character-level language modeling}
The final task we consider is character-level language modeling. The goal is to predict the next character in a sequence of characters. To that end, we stack up to eight QRNN layers, either using DReLUs or DELUs, and compare our results with the current state-of-the-art.

\subsubsection{Experimental setup}
For our character-level language modeling experiments, we consider a small and large more challenging dataset. The small dataset is again the Penn Treebank (PTB) dataset \citep{ptb} consisting of roughly 6M characters, while the large dataset is the \emph{enwik8}/Hutter Prize dataset \citep{hutter} which contains 100M characters extracted from Wikipedia. This dataset is challenging because it contains XML markup and non-Latin characters.  We adopt the same train/validation/test split as \cite{mikolov2012subword} for the PTB dataset, and the 90M/5M/5M dataset split for the Hutter Prize dataset.

Our neural network architecture consists of an embedding layer, a number of QRNN layers, and a final classification layer with a softmax activation function. We use embeddings of size 50 and hidden states of size 250, 500, and 1000. The number of QRNN layers is 2, 4, or 8. The width $n$ of the convolution of the first layer is always six, while at the other layers the convolution width $n$ is two. The weight matrices are orthogonally initialized. We trained the neural network using Adam \citep{kingma2014adam} with learning rate 0.0003 and constrained the norm of the gradient at 5. Following \cite{hypernetworks}, we used a batch size of 128 and a sequence length of 100 for the language modeling experiments on the Penn Treebank dataset. Due to memory constraints, we used the same batch size of 128 for experiments on the Hutter Prize dataset, but we used a sequence length of 200 instead of 250. We regularized the model using dropout on the output of every layer \citep{Zaremba} with a dropout probability of 0.15 for models with hidden state size 250 and a dropout probability 0.3 for models with hidden state size 500 and 1000. We applied batch normalization in all QRNN models \citep{conf/icml/IoffeS15,DBLP:journals/corr/CooijmansBLC16}. The evaluation metric used is Bits-Per-Character (BPC).



\subsubsection{Discussion}

\paragraph{Penn Treebank experiments}

In Table~\ref{tab:tab_char1}, the results of several experiments on the PTB dataset are listed. The first part shows the results of a number of successful and state-of-the-art character-level language modeling methods, including a vanilla two-layer LSTM, while the second part shows our results for QRNNs trained with several activation functions.

In general, we can observe that the BPC score of \emph{tanh}-based QRNNs is worse than the BPC score of DReLU- or DELU-based QRNNs. DReLUs perform equally good as DELUs. Doubling the number of QRNN layers from four to eight is more parameter efficient than doubling the hidden state size from 250 units to 500 units. The language modeling performance, however, is the same. 
When using eight QRNN layers and 500 hidden units, we obtain a BPC score of 1.21, outperforming a two-layer HyperLSTM with a similar number of parameters, a new state-of-the-art result on this dataset. 


\paragraph{Hutter Prize experiments}
Compared to the PTB dataset, the Hutter Prize dataset is a more challenging dataset consisting of 100M characters, as well as non-Latin characters and XML markup. The results of our experiments are depicted in Table~\ref{tab:tab_char2}. 
Apart from the number of layers, we trained two identical QRNN-based architectures. We only consider DReLUs, given that DELUs gave similar results on all other benchmarks. Our model with four layers and 1000 hidden state units in each layer obtains a BPC score of 1.32 and performs better than HyperLSTMs using a similar number of parameters (25.82M) and equally good as Hierarchical Multiscale LSTMs (HM-LSTM) \citep{chung+al-2017-multiscale-iclr}. When doubling the number of layers to eight, we further reduce the BPC score to 1.25, outperforming more complicated architectures such as ByteNet \citep{DBLP:journals/corr/KalchbrennerESO16} and Recurrent Highway Networks \citep{DBLP:journals/corr/ZillySKS16}. Note that the latter was only evaluated in the context of language modeling. 

\section{Conclusions}\label{sec:conclusions}
In this paper, we introduced Dual Rectified Linear Units (DReLUs) and the exponential extension Dual Exponential Units (DELUs), demonstrating that they are valid drop-in replacements for the \emph{tanh} activation function in Quasi-Recurrent Neural Networks (QRNNs). Similar to \emph{tanh} units, DReLUs have both positive and negative activations. In addition, DReLUs have several advantages over \emph{tanh} units: (1) they do not decrease the magnitude of gradients when active, (2) they can be exactly zero, making them noise robust, and (3) they induce sparse output activations in each layer. 

We evaluated DReLUs and DELUs on three different tasks. We evaluated on sentiment classification and showed that DReLUs and DELUs do not need dense connections to improve gradient backpropagation compared to \emph{tanh} activation functions when stacking four layers of QRNNs. We also demonstrated that DReLUs and DELUs improve perplexity on the task of word-level language modeling compared to \emph{tanh} activation functions and that a single ReLU is an inferior replacement. Finally, we trained a model with eight stacked DReLU- and DELU-based QRNN layers and obtained state-of-the-art results for the task of character-level language modeling on two different datasets, over \emph{tanh}-based QRNNs and more advanced architectures \citep{DBLP:journals/corr/ZillySKS16,DBLP:journals/corr/KalchbrennerESO16}.

\section*{Acknowledgments}


The research activities as described in this paper were funded by Ghent University, imec, Flanders Innovation \& Entrepreneurship (VLAIO), the Fund for Scientific Research-Flanders (FWO-Flanders), and the EU.
\bibliography{mybibfile}

\end{document}